# Transmission Line Detection Based on Improved Hough Transform

Wei Song, Pei Li, Man Wang

School of Information Science and Engineering, Chongqing Jiaotong University

**Abstract**

To address the challenges of low detection accuracy and high false positive rates of transmission lines in UAV (Unmanned Aerial Vehicle) images, we explore the linear features and spatial distribution. We introduce an enhanced stochastic Hough transform technique tailored for detecting transmission lines in complex backgrounds. By employing the Hessian matrix for initial preprocessing of transmission lines, and utilizing boundary search and pixel row segmentation, our approach distinguishes transmission line areas from the background. We significantly reduce both false positives and missed detections, thereby improving the accuracy of transmission line identification. Experiments demonstrate that our method not only processes images more rapidly, but also yields superior detection results compared to conventional and random Hough transform methods.

## 1    Introduction

With the continuous expansion of the investment scale of the national power grid, the network structure is becoming more and more complex, and the workload of power grid inspection and maintenance is huge [1], the traditional manual inspection and operation of transmission lines and substations can no longer meet the requirements of efficient power grid inspection work. Therefore, the State Grid Corporation has vigorously promoted UAV line inspection [2-5]. Since the background of aerial images is a complex and variable natural background, direct inspection of transmission lines with aerial images will result in a high false detection rate and leakage rate.

In order to reduce the interference of background noise on transmission line targets, many methods have been studied in recent years in China for detecting transmission lines in aerial images. Zhao Lipo et al. [6] used orientation constraint-based linear target enhancement for transmission line targets to inhibit the interference objects in the straight direction and other nonlinear backgrounds and noises, and introduced a recognition factor by Radon transform to remove the horizontal interfering objects, but this method has harsh constraints and can only recognize the horizontal interfering objects, which can only be recognized with a high accuracy. However, this method is very restrictive, and it can only recognize transmission lines in approximately horizontal direction, which is very limited. Cao Wei et al. [7] used autocorrelation enhancement with directional filtering to enhance the power line targets while greatly reducing the complex environmental background in aerial images, which effectively improves the detection and recognition rate of power line targets in images. However, the efficiency of the algorithm and the enhancement effect depend on the number of iterations, and it is necessary to manually control the number of iterations to achieve the best effect. Huang, Dongfang et al. [8] used search clustering to extract pixels and combined with morphological processing pixels to remove background noise from images. An adaptive estimation method of threshold interval is used to calculate the parameter threshold of the Hough transform to identify the transmission lines in the image. However, the selection process of parameter thresholds is complicated, time-consuming, and the recognition effect is poor in the case of low contrast.

In view of the shortcomings of these methods to deal with the complex background environment, this paper is based on the previous research in transmission line detection, combined with a large number of aerial image examples and transmission line features to design a simplified model of the transmission line in aerial images, determine the boundary of the transmission line region, the transmission line region and get the segmentation coefficients. Eliminate the background noise of the non-transmission line area, and calculate the distance between the lines in the transmission line area to delete the non-compliant lines, reduce the time required for traversing the parameter space in the stochastic Hough transform, and reduce the probability of misdetection and loss of transmission line targets.

## 2	Image preprocessing

To analyze and process aerial images, there are requirements for accuracy, real-time, robustness. In the analysis of aerial images must be pre-processed before, so as to enhance the characteristics of the transmission line, reduce the purpose of background interference.

The most common methods for image preprocessing are Canny algorithm edge detection method [9], Canny method has higher detection accuracy and more application scenarios [10], due to Canny algorithm is more sensitive to the noise, it is not suitable for the edge detection of transmission line under the complex background. The Hessian matrix method [11, 12] has been widely used in the enhancement of linear targets in images, compared with the Canny method, this paper uses the Hessian method to enhance the aerial images and extract all the edge points.

$$H(f)(x,y) = \begin{bmatrix} \dfrac{\partial^2 f}{\partial x^2} & \dfrac{\partial^2 f}{\partial x \partial y} \\ \dfrac{\partial^2 f}{\partial y \partial x} & \dfrac{\partial^2 f}{\partial y^2} \end{bmatrix}$$

The Hessian matrix describes the changes of the gray gradient in different directions, and the eigenvalues and eigenvectors of the corresponding pixels are obtained from the Hessian matrix.

## 3	Improved Stochastic Hough Transform Transmission Line Detection Methods

### 3.1	Standard Hough Transform and Random Hough Transform

The standard Hough transform is defined in [13], and its basic idea is to use point-line duality to transform the overall detection problem into a local detection problem.

To avoid the problem of infinite slope of a vertical line, the polar equation of the line is usually used

$$\rho = x\cos\theta + y\sin\theta$$

Where: $\rho$ is the distance from the origin of the right-angle coordinate system to the line, $\theta$ is the angle between the line and the x-axis.

Figure 1 has a set of points on the same line, each point corresponds to a sine curve in the parameter space, these curves have a common intersection in the polar coordinate system. Set up an accumulator set $A[\theta, \rho]$ with initial value 0, and for each pixel $(x_i, y_i)$, iterate through the values of $\theta$, calculate the corresponding $\rho$. If the value corresponds to one of the values of the accumulator set, the accumulator will be increased by 1 until all the pixels have been iterated through.

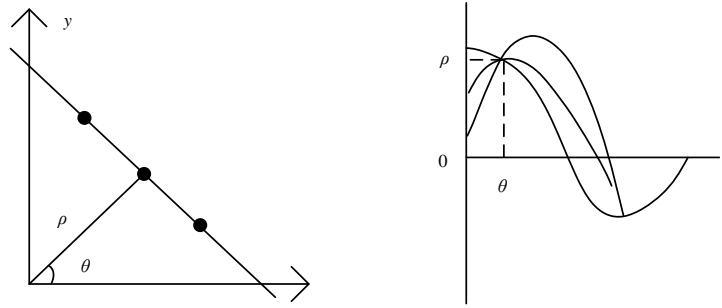

**Fig. 1 Point-line pairing in polar coordinates.**

Since the standard Hough transform maps every pixel in the image, the computational complexity is high, which is not suitable for the processing of UAV aerial images. The basic idea of stochastic Hough transform based on probabilistic statistics can be found in the literature [14, 15], which avoids the huge computation required by the one-to-many mapping in the standard Hough transform by using the many-to-one mapping, but there is still the problem of invalid accumulation, and it is easy to detect spurious transmission lines or omission of transmission lines in complex scenes.

After our statistical analysis of a large number of UAV aerial images, more than 80% of the aerial images of transmission lines only account for less than 40% of the whole area. However, the random Hough transform does not take into account the fact that transmission lines will be concentrated in one area in transmission line detection, and randomly picking up points in the whole image space will waste a lot of time and introduce background noise in non-transmission line areas.

### 3.2  Improved Randomized Hough Transform

In this paper, we improve the random Hough transform method and propose a random Hough transform method based on region segmentation. The main idea is to mark the boundary of the transmission line region and divide it into transmission line regions according to the regional distribution of transmission lines in the image, and introduce a slope judgment mechanism to reduce the false detection caused by the false peak detection through the parallel and pairwise appearance of transmission lines, so as to improve the speed and effect of transmission line detection.

By statistically analyzing a large number of drone aerial images, it is found that more than 60% of the aerial images have the following characteristics.  (1) the topology of transmission lines in aerial images is straight and passes through the images; (2) the transmission lines are parallel to each other, and the distance between them is fixed; (3) the width of transmission lines is about 1 to 2 pixels; and (4) the transmission lines usually appear in pairs. Based on the four characteristics described above, a simplified model $S$ can be designed for the transmission lines in aerial images. The model can be applied if the transmission lines in aerial images meet the above four characteristics. As shown in Fig. 2, $L_1, L_2, L_3, L_4$ are four transmission lines, and the distances between them are $d_1, d_2$, and $abcd$ are the endpoints of the transmission line area boundary, and the transmission line area $S'$ is the quadrilateral formed by $abcd$ as the vertex.

The next section describes the transmission line region delineation and model generation process. The Hessian preprocessed aerial image has detected all the edge information in the image, including the transmission line targets. Since the transmission line always crosses from one end of the image to the other, the search starts from the boundary of the image, and the boundary is the starting point of the search. The following three steps are used to delineate the transmission line area:

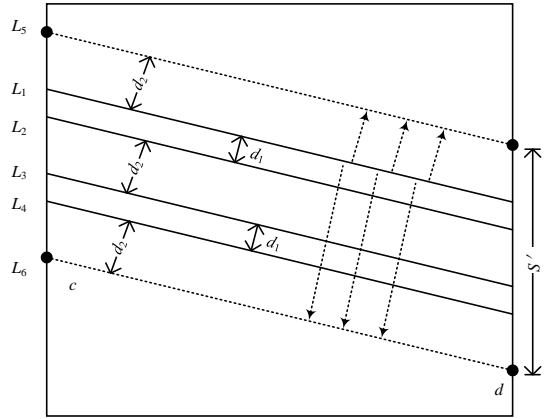

**Fig. 2 Simplified model of the transmission line in the aerial image S**

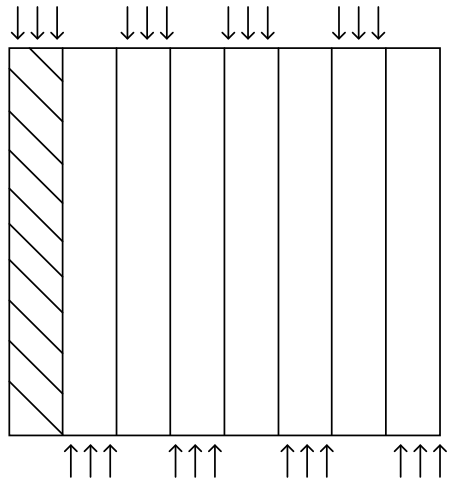

**Fig. 3 Pixel bar segmentation for transmission line area delineation.**

(1) First, search from the top to the bottom, first vertically divide the whole image into 8 blocks from left to right, forming 8 pixel strips with the same width, as shown in Fig. 3, and then search vertically downward with the upper boundary as the starting point. Then we search for the edges of the line segments within this pixel strip, and use the sub-linear algorithm [17] to find the first longest edge of the line, which is part of $L_1$, and then continue to search downward to find $L_2, L_3$, and $L_4$ segments in turn. The maximum spacing between segments is $d_2$ and the minimum spacing is $d_1$. After finding the $L_4$ fragment, the search continues to the bottom border of the image, and no fragments are found at a distance greater than $d_2$, which is where the dotted line $L_6$ is located. The position of the dotted line $L_6$ is labeled as the lower boundary of the transmission line area in the first pixel bar.

(2) Next, search from the bottom to the top, taking the lower boundary as the starting point, and start searching vertically upwards. The first segment found is the $L_4$ segment, followed by the $L_2, L_3$, and $L_1$ segments. After finding the $L_1$ fragment, the search continues up to the upper image boundary, where it is also found that there are no line segments larger than the $d_2$ spacing, i.e., where the dotted line $L_5$ is located. The location of the dotted line $L_5$ is labeled as the upper boundary of the transmission line area in the second pixel bar.

(3) Splice the upper and lower boundaries of the transmission line region of all pixel strips to get the complete transmission line region $S'$ surrounded by the four vertices of $abcd$.

After dividing the transmission line area, we can get the values of $d_1, d_2$ and the related parameters of the transmission line area, thus forming a specific simplified model that conforms to the image.

Setting the area segmentation coefficient $I_c$, $I_c$ is the ratio of the sum of pixels in the transmission line area to the sum of pixels in the whole image; $N'$ is the number of pixels in the image after segmentation, $N$ is the number of pixels in the image before segmentation, si and $S'_j$ correspond to the pixels in the image before and after segmentation, respectively.

Regional segmentation coefficient

$$I_c = \frac{\sum_{j=0}^{j=N'} S'_j}{\sum_{i=0}^{i=N} S_i}$$

Considering that a power transmission line consists of $n$ pixels, the probability of detecting the power transmission line in the parameter space image is $P_c$ in an experiment of randomly sampling two pixels;

Random Hough transform method

$$P_c = \frac{n(n-1)}{N(N-1)}$$

Improved method in this paper

$$P_c = \frac{n(n-1)}{(N*I_c)(N*I_c-1)}$$

Due to the region splitting coefficient $0 < I_c < 1$, the $P_c$ obtained by the improved method is larger than that by the traditional method.

After $M$ experiments, the number of times the transmission line is detected is a variable in the binomial distribution $\xi$

$$P_c(\xi = k) = C_M^k P_c^k (1 - p_c)^{M-k}$$

Then the probability of losing the transmission line in $k_0$ experiments is $P_{miss}$

$$P_{miss} = \sum_{k=0}^{k_0} P_c(\xi = k)$$

As $P_c$ increases, $P_c(\xi = k)$ becomes smaller, so Pmiss becomes relatively smaller.

Consider a mis-detected transmission line consisting of $m$ pixels, and the The probability of detecting the transmission line in the parameter space image is $P_r$;

Random Hough transform method

$$P_r = \frac{m(m-1)}{N(N-1)}$$

Improved method in this paper

$$P_r = \frac{(m * I_c)(m * I_c - 1)}{(N * I_c)(N * I_c - 1)}$$

After $M$ experiments, the number of times the power line is detected is a variable $\xi$ in binomial distribution.

$$P_r(\xi = k) = C_M^k P_r^k (1 - P_r)^{M-k}$$

Then the probability of false detection of the transmission line in $k_0$ experiments is $P_{\text{false}}$

$$P_{\text{false}} = 1 - \sum_{k=0}^{k_0} P_r(\xi = k)$$

As $P_r$ increases, $P_r(\xi = k)$ decreases, so $P_{\text{false}}$ is relatively small.

In this paper, we firstly constructed a simplified model diagram of aerial transmission line in combination with the practical situation, and determined the boundary of transmission line area through the description of the model diagram to limit the transmission line area. By calculating the region splitting coefficient, a large amount of irrelevant transmission line information background noise is separated from the aerial image, thus greatly reducing the number of traversing pixels in the parameter space, improving the speed of transmission line detection, and reducing the probability of loss and false detection of the transmission line target, and improving the probability of detection of the transmission line target.

### 3.3 Transmission line detection algorithm

In this paper, we use the improved stochastic Hough method to process the aerial images, although it reduces the interference caused by the background noise in the non-transmission line area, but there is still the problem of false peak detection.

If the national value is set to 88, if there are three cumulative units with a national value greater than 88, these three cumulative units will be detected as three transmission lines, resulting in false detection. In this paper, by utilizing the characteristics of transmission lines appearing in pairs and in parallel, the slope comparison of the detected straight lines and the distance between the straight lines can be calculated and processed, eliminating the transmission lines that do not meet the requirements, thus greatly reducing the false detection of transmission lines due to false peak detection.

Take any two points $(x_m, y_m)$ and $(x_n, y_n)$ on the line, where

$$k = \frac{y_m - y_n}{x_m - x_n}$$

Any two parallel straight lines $Ax + By + C_1 = 0$ and $Ax + By + C_2 = 0$, the distance.

$$d_i = \frac{|C_1 - C_2|}{\sqrt{A^2 + B^2}}$$

Transmission line detection algorithm flow is as follows:

Input: binary image of aerial image.

Output: a binary image of the marked power line.

Step 1, establishing an image space and a parameter space;

Step 2 Determine the boundaries of the transmission line region, calculate the region splitting coefficients, and divide the transmission line region and the irrelevant region; Step 3 Determine the parameterization of the transmission line region, and calculate the region splitting coefficient.

Step 3 Discretize the parameter space of the transmission line region into an accumulator array, and map each point in the image to the accumulator corresponding to the parameter space.

Step 4: Detect the local maxima of the accumulator array, obtain the corresponding parameters, and retain all the candidate transmission lines by a preset threshold; Step 5: Calculate all the candidate transmission lines.

Step 5 Calculate the slopes of all candidate transmission lines, count the different slope values, and retain the transmission line with the highest count; Step 6 Calculate the slopes between parallel transmission lines.

Step 6 Calculate the mutual distance between parallel transmission lines, and the variance of any one of the four values in Fig. 2 is less than 0.01, then the transmission line is retained.

## 4    Experimental results and analysis

In order to verify the image detection effect and time required for detecting aerial transmission lines based on the improved stochastic Hough transform proposed in this paper, the experimental platform is Win 10 operating system. The experimental platform is Win 10 operating system, Visual Studio 2012 + OpenCV 2.4.13. In this paper, three groups of UAV aerial images are selected, and firstly, the Canny algorithm edge detection method and the Hessian method are used to preprocess the three groups of images, and then the standard Hough transform algorithm, the stochastic Hough transform algorithm and the improved stochastic Hough transform method are used to detect the transmission lines. Then the standard Hough transform algorithm, randomized Hough transform algorithm and the improved randomized Hough transform method in this paper are used to compare the detection effect and the time required, respectively.

## 4.1 Comparison of preprocessing effect of two methods

In this experiment, three sets of UAV aerial images of transmission lines with different complex backgrounds are used, and the image resolution is 620 ∗ 810. The Canny method and Hessian method are applied to the aerial images respectively, and the experimental effects are shown in Fig. 4. The experimental results are shown in Fig. 4. From the whole effect of the three groups in Fig. 4, the Canny method detects a large number of background edge lines, and part of the transmission line is covered by the edge lines, which affects the detection effect.

Using the Hessian method to enhance the experimental images, the first group and the third group get better edge points, the enhancement effect of the second group is not as good as that of the first group and the third group because of the most complicated background and the interference of fields in the background. However, the linear features of power lines are enhanced, and the complex natural background is smoothed to remove the interference information of some forest trees, which is beneficial to the next algorithm to detect power lines in the image.

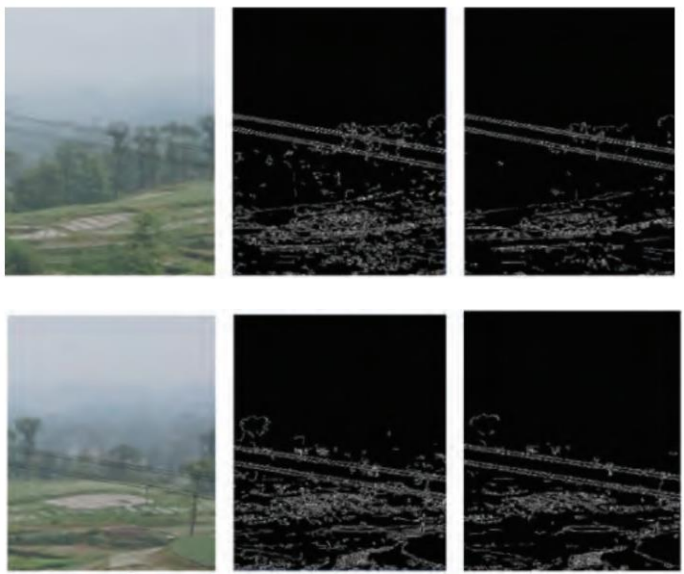

Fig. 4 Edge detection effect of Canny method and Hessian method.

## 5 Comparison of detection effect and performance of 23 transmission line detection algorithms

After the pre-processing of the three groups of experimental images by the Hessian method, the transmission line target is enhanced to a certain extent. Based on the preprocessing, the standard Hough transform, random Hough transform, and the improved random Hough transform are applied to the three groups of images, and the experimental results are shown in Fig. 5.

In Fig. 5, there are more fields and trees in the first group, and the traditional Hough transform detects a lot of background noises when detecting transmission lines, especially the blurring of the edges of the transmission lines where the transmission line targets overlap with the forest background, and the stochastic Hough transform reduces the noises of the fields near the transmission lines to a certain degree, but it still affects the overall detection effect, and the method in this paper filters the noise of the transmission lines very well. In this paper, the background noise near the transmission line is well filtered, and the extracted line segments are coherent and complete, and the overall detection effect is better than that of the standard Hough transform and the random Hough transform. In the second group, compared with the first and third groups, the background noise is the most complicated, no matter it is the traditional Hough transform or random Hough transform, a lot of interfering information is detected in the field and the dome of the hut, and some of the interfering information is mixed with the transmission line, which seriously affects the detection effect of the transmission line, and there are many breaks on the transmission line, and the information of the dome and the field is weakened in the present method, so as to inhibit the background noise, although the transmission line section is more coherent than the standard Hough transform and random Hough transform. In this method, the dome and field information are weakened to suppress the background noise, and although the clarity of the transmission line target is somewhat reduced, the transmission line target is still clearly recognized, and the detection effect is better than that of the standard Hough transform and the random Hough transform in general. In the third group, the tail of the transmission line is seriously interfered by the field soil information, and the standard Hough transform and random Hough transform detect the tail and the background field soil information that interfere with each other, resulting in incomplete edge information, which makes it impossible to identify the transmission line, but in this paper, the edge of the tail of the transmission line is separated from the field soil and field information is weakened, which results in the detection of the tail end not as good as the other parts of the transmission line. Although the detection effect of the tail end is not as good as that of other parts of the transmission line, the overall detection effect is still better than that of the standard Hough transform and the random Hough transform.

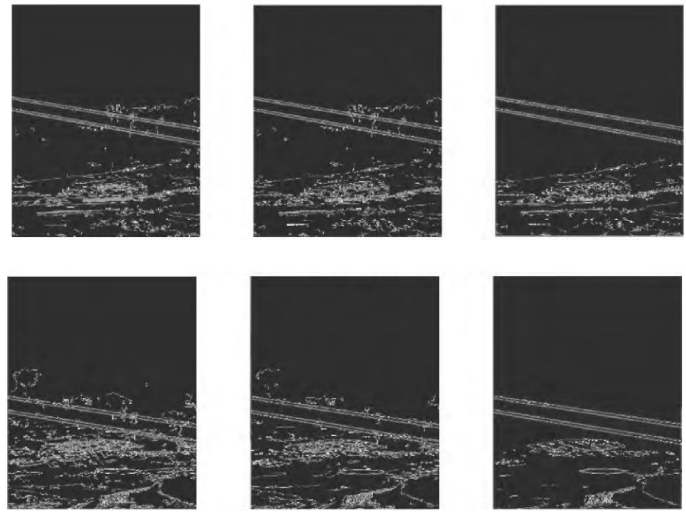

**Fig. 5 Comparison between the traditional method and the method of this paper.**

The hardware platform used in this experiment is as follows. The CPU is i7-3520M, main frequency 2.9GHZ, memory 8 GB, and the performance is evaluated by the time required for detecting a straight line in the three methods. Table 1 lists the time consumed by the three different methods in detecting the transmission line and the thresholds in detecting the transmission line. In the case of close thresholds, in group a, the detection time of the transmission line by the proposed method is 35% of the standard Hough transform and 58% of the random Hough transform. b In group c, the detection time of the transmission line by the proposed method is 34% of the standard Hough transform and 58% of the random Hough transform. c In group d, the detection time of the transmission line by the proposed method is 34% of the standard Hough transform and 58% of the random Hough transform. d In group e, the detection time of the transmission line by the proposed method is 34% of the standard Hough transform and 58% of the random Hough transform.   In group c, this method is 34% better than the standard Hough transform and 58% better than the random Hough transform. This shows that the method proposed in this paper can effectively reduce the time required for power line detection and improve the performance of power line detection in aerial images with complex background.

**Table 1 Comparison of experimental results of three methods.**

| Experiment Group | Standard Hough transform | | Random Hough transform | | Method | |
|---|---|---|---|---|---|---|
| | Threshold | Elapsed Time /ms | Threshold | Elapsed Time /ms | Threshold | Elapsed Time /ms |
| a | 112 | 200.8 | 113 | 123.0 | 113.0 | 72.2 |
| b | 115 | 210.0 | 114 | 132.1 | 112 | 81.8 |
| c | 116 | 202.2 | 113 | 120.3 | 115 | 70.2 |

**6    Conclusion**

Aiming at the problems of time-consuming detection of transmission lines by the traditional Hough transform method and the high probability of loss and false detection of transmission lines by the stochastic Hough transform method, an improved stochastic Hough transform method based on area segmentation is proposed, in which the transmission line area and non-transmission line area are separated by the area segmentation coefficients to reduce the interference of the background noise, and then the effects of false peaks are reduced by the spacing of the transmission lines. The experimental results show that the proposed method is suitable for UAVs to inspect transmission lines in the natural environment of complex forest vegetation in the field, and it can not only detect transmission lines in aerial images more accurately, but also reduce the detection time and detect transmission lines in aerial images quickly and effectively, which is practical to a certain extent.